\documentclass[lettersize,journal]{IEEEtran}
\usepackage{amsmath,amsfonts}
\usepackage{algorithmic}
\usepackage{algorithm}
\usepackage{array}
\usepackage[caption=false,font=footnotesize,labelfont=rm,textfont=rm]{subfig}
\usepackage{textcomp}
\usepackage{stfloats}
\usepackage{url}
\usepackage{verbatim}
\usepackage{graphicx}
\usepackage{cite}
\usepackage{color}
\usepackage{svg}
\usepackage{makecell}
\usepackage{multirow}
\begin{document}

\title{Prototype-Guided Text-based Person Search based on Rich Chinese Descriptions}

\author{Ziqiang Wu, 
        Bingpeng Ma
\thanks{Z. Wu and B. Ma are with University of Chinese Academy of Sciences, Beijing 100049, China.}
\thanks{E-mail: wuziqiang19@mails.ucas.ac.cn, bpma@ucas.ac.cn}
}

\markboth{Journal of \LaTeX\ Class Files,~Vol.~14, No.~8, August~2021}%
{Shell \MakeLowercase{\textit{et al.}}: A Sample Article Using IEEEtran.cls for IEEE Journals}


\maketitle

\begin{abstract}
Text-based person search aims to simultaneously localize and identify the target person based on query text from uncropped scene images, which can be regarded as the unified task of person detection and text-based person retrieval task. In this work, we propose a large-scale benchmark dataset named PRW-TPS-CN based on the widely used person search dataset PRW. 
Our dataset contains 47,102 sentences, which means there is quite more information than existing dataset. These texts precisely describe the person images from top to bottom, which in line with the natural description order. We also provide both Chinese and English descriptions in our dataset for more comprehensive evaluation. These characteristics make our dataset more applicable. 
To alleviate the inconsistency between person detection and text-based person retrieval, we take advantage of the rich texts in PRW-TPS-CN dataset. We propose to aggregate multiple texts as text prototypes to maintain the prominent text features of a person, which can better reflect the whole character of a person. The overall prototypes lead to generating the image attention map to eliminate the detection misalignment causing the decrease of text-based person retrieval. Thus, the inconsistency between person detection and text-based person retrieval is largely alleviated.
We conduct extensive experiments on the PRW-TPS-CN dataset. The experimental results show the PRW-TPS-CN dataset's effectiveness and the state-of-the-art performance of our approach.
\end{abstract}


\section{Introduction}
\IEEEPARstart{P}{erson} re-identification(Re-ID) plays an indispensable role in intelligent video surveillance and is an important field in computer vision. Due to the limitation of image-to-image person retrieval, there are lots of cross-modal person retrieval methods have been proposed, including Infrared-RGB Person Re-ID\cite{choi2020hi, lin2022learning, zhang2022fmcnet}, Text-based Person Retrieval (TPR)\cite{wang2020vitaa, niu2020improving, farooq2021axm, chen2021cmka} , Text-based Person Search (TPS)\cite{zhang2021text}, Sketch-RGB Person Re-ID\cite{yang2019person, chen2022cross}. Text-based person search task searches the target person in original surveillance images by using textual descriptions. This technology can be used more effectively in nuisance tracking and missing persons finding.

\begin{figure}[!t]
\centering
    \subfloat[Text-based Person Retrieval]{
		\includegraphics[width=\linewidth]{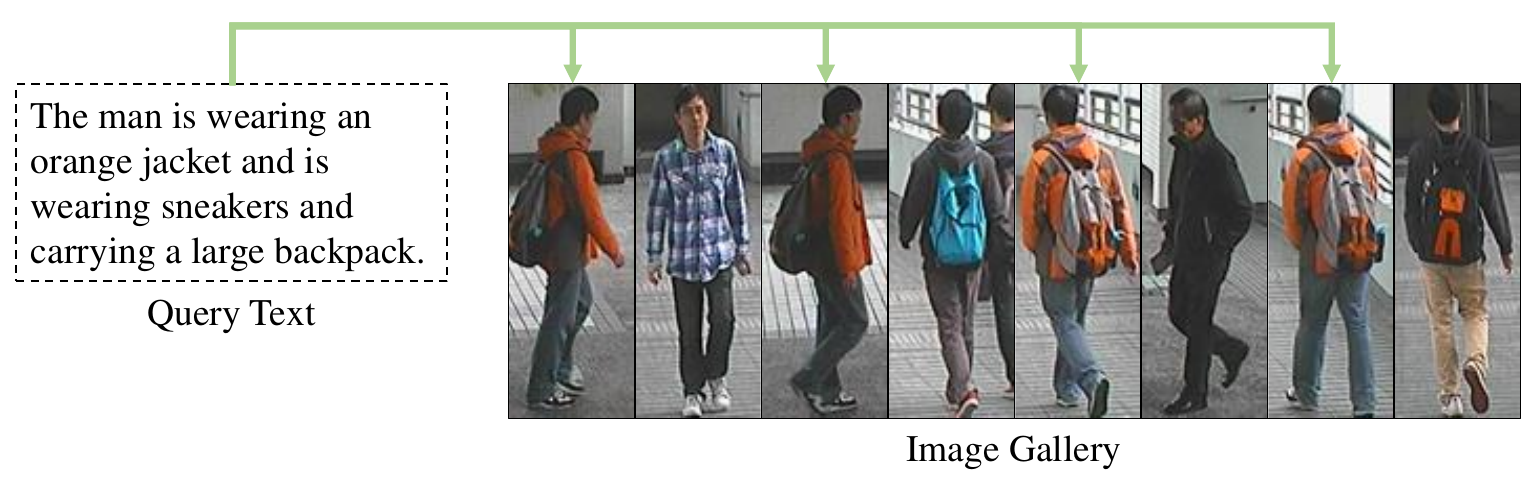}	
    }
    \hfill
    \subfloat[Text-based Person Search]{
		\includegraphics[width=\linewidth]{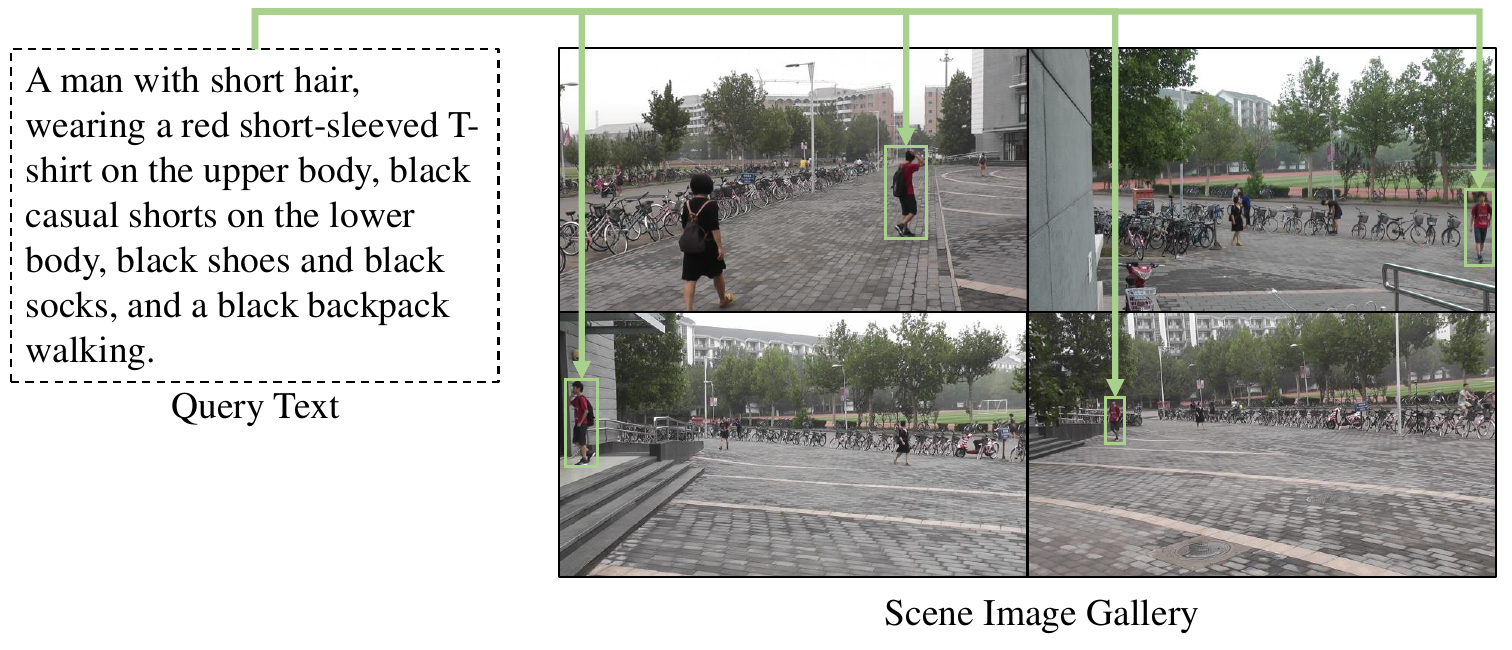}	
    }
\caption{Comparison of two types of text-based person re-identification tasks: (a) Text-based Person Retrieval, using text to retrieve cropped person images from the image gallery. (b) Text-based Person Search, using text to search person from the uncropped scene image gallery.}
\label{Illustration of tasks}	
\end{figure}

In this paper, we aim at the text-based person search task. The comparison between text-based person search and text-based person retrieval is shown in Figure. \ref{Illustration of tasks}. Compared with text-based person retrieval, text-based person search is supposed to detect persons in original surveillance images by using texts, which is more challenging and applicable. 
It makes up for the impracticality of the traditional Person Re-ID task that needs to manually crop the image, and break the limitation of person search that needs the target person image in advance.

Some researchers have recognized the importance of the task and provided dataset and method\cite{zhang2021text}. However, both the dataset and method are inadequate. For example, the training data in the dataset is insufficient. Each training cropped image, is only labeled with one description. Lack of training text will make the model easy to learn the biases of the annotators, causing insufficient training. Compared with one training description in \cite{zhang2021text}, the descriptions number of each image are generally equal or larger than 2 in the wide-used dataset, such as 2 in CUHK-PEDES\cite{li2017person}, 5 in Flickr30K\cite{young2014flickr30k}, 5 in MS-COCO\cite{lin2014microsoft}, 10 in CUB\cite{reed2016learning}, 10 in Flowers\cite{reed2016learning}. Thus, a dataset with multiple descriptions is desperately demanded to facilitate this task.


As for the method, previous research directly uses shared image features for person detection and text-based person retrieval. Person detection aims to find the common features of human, but text-based person retrieval aims to find the differences among identities. This conflict causes feature ambiguity for text-based person retrieval. Considering the ambiguity, it is also a good idea to maintain the two different stages and adapt the two stages to each other. 

To solve the mentioned problems, we first propose a text-based person search dataset named PRW-TPS-CN. There are a total of 47,102 descriptions including 17,310 queries in the dataset, containing more information than the existing dataset \cite{zhang2021text}. Considering that existing methods \cite{li2017person, young2014flickr30k, lin2014microsoft, reed2016learning} show that more annotated sentences will largely improve the performance, we provide 2 Chinese descriptions for each training person in bounding box. The descriptions are informative and described in the top-to-bottom order of images. These structured sentences help the model learn the features more easily. Besides, we also translate Chinese descriptions into English and provide the English version of the dataset, which makes the dataset more applicable. Our dataset can better cope with the challenges of text-based person search, thereby facilitating the development of this problem.

Furthermore, we propose a network containing Text Prototype Attention Network (TPAN) to significantly improve the practicability of text-based person search. 
To alleviate the inconsistency between person detection and text-based person retrieval, we take advantage of the rich texts in the PRW-TPS-CN dataset. Considering that the feature prototypes can maintain the salient features which can largely guide the generation of image attention, we embed a prototype table in our proposed network. In detail, we aggregate all the textual features of the same person as the prototype and maintain and update them in a prototype table. These overall prototypes are used to guide the generation of person image attention map instead of query-driven in \cite{zhang2021text}. With the attention maps, the common features of the same person in different detected images will be highlighted, which will largely eliminate the detection misalignment.

We verify the effectiveness of the PRW-TPS-CN dataset by implementing the baseline model and achieving 44.90\% in Rank-1 and 27.71\% in mAP, which proves the dataset is feasible. With TPAN, we surpass the baseline by 1.29\% and 1.23\% in mAP and Rank-1 in the PRW-TPS-CN dataset. This improvement shows the proposed TPAN largely alleviates the feature ambiguity and misalignment problem.

\section{Related Work}
\subsection{\bf{Text-based Person Retrieval}}

In similarity learning, the network learns the similarity between images and texts based on fine-grained partitions. Niu \textit{et al.} \cite{niu2020improving} align image and text features by similarity calculation in three granularities, which is named Multi-granularity Image-text Alignment (MIA). Wang \textit{et al.} \cite{wang2020vitaa} propose the Visual-Textual Attribute Alignment (ViTAA) to disentangle image and text into attribute feature space and calculate the similarity. Farooq \textit{et al.} \cite{farooq2021axm} propose Aligned Cross-Modal Net (AXM-Net) to learn the cross-modal semantic feature alignment. These methods tend to be time consuming.

For implicit relation reasoning, Jiang \textit{et al.} propose IRRA\cite{cvpr23crossmodal} to learn relations between local visual-textual tokens and enhances global image-text matching without requiring additional prior supervision. This requires the representation ability of model large enough.

Common space learning aims at projecting the image and text into a common space to eliminate the modal gap. Sarafianos \textit{et al.} \cite{sarafianos2019adversarial} propose the Text-Image Modality Adversarial Matching (TIMAM) to project both modalities into a common space based on the idea of Generative Adversarial Network (GAN). Zheng \textit{et al.} \cite{zheng2020dual} propose a dual-path network, and use instance loss to learn more fine-grained modal features in a common space. By considering text knowledge is limited, Wu \textit{et al.} \cite{Refined} propose RKT to transfer the knowledge from image to text and further refine the knowledge.
These methods are devoted to eliminating the modal gap by using global features. Due to that common space learning is faster and more practical, we adopt it in our method.

\subsection{\bf{Person Search}}
The person search task aims to locate and identify a person simultaneously. The most direct way is to combine the person detector with the person Re-ID modules. In 2014, Xu \textit{et al.} \cite{xu2014person} first introduce the concept of person search and use the hand-craft method to extract the person features. According to whether sharing the person features in the two modules, person search can be divided into two categories: two-stage methods\cite{zheng2017person, chen2018person, lan2018person, han2019re, dong2020instance, wang2020tcts} and one-stage methods\cite{xiao2017oim, liu2017neural, chang2018rcaa, xiao2019ian, yan2019learning, munjal2019query, chen2020norm, kim2021prototype, han2021end, yan2021alignps, cao2022pstr, yu2022cascade}. 

One-stage methods aim to solve the person search end-to-end, which means the person detector and Re-ID modules share the same person features. Xiao \textit{et al.} \cite{xiao2017oim} first introduce an end-to-end person search model and train the person detector and Re-ID models jointly. Yan \textit{et al.} \cite{yan2021alignps} propose the first anchor-free person search framework in the principle of ``re-id first''. With the development of transformer structure, Cao \textit{et al.} \cite{cao2022pstr} and Yu \textit{et al.} \cite{yu2022cascade} propose the transformer-based person search frameworks.

Two-stage methods solve the person search task in two steps, including training an independent person detector and a Re-ID model. Zheng \textit{et al.} \cite{zheng2017person} propose the person search dataset Person Re-identification in the Wild (PRW) and evaluate multiple combinations of different person detector and Re-ID models. Han \textit{et al.} \cite{han2019re} propose a method named Localization Refinement (LR) for providing the refined detection boxes for person search. To make 
person detection and Re-ID module more consistent, Wang \textit{et al.} \cite{wang2020tcts} propose IDentity-Guided Query (IDGQ) and Detection Results Adapted (DRA) modules, making adaptations to both steps. The advantage of the two-stage methods is to extract features from different stages separately, thus the features can aim at their own task. By considering the advantage, we adopt two-stage framework as our base network.

\subsection{\bf{Prototype}}
Prototype represents training examples for better feature extraction, also known as proxy\cite{movshovitz2017no, qian2019softtriple}. Unlike traditional deep learning methods focusing on instance features, prototype-guided methods optimize features to be close to their prototype. The prototype is widely used in deep metric learning to minimize intra-class variation.

Prototype can play different roles in network training. Traditionally, Xiao \textit{et al.} \cite{xiao2017joint} propose an Online Instance Matching (OIM) method to keep the prototype of each identity for training the network effectively. Besides, Wang \textit{et al.} \cite{wang2020cross} propose the Cross-Batch Memory (XBM), which memorizes the embeddings of past iterations and allows the model to collect sufficient hard negative pairs. Kim \textit{et al.} \cite{kim2021prototype} propose Prototype-Guided Attention (PGA) to maintain the image prototypes for generating attentions. Considering the large intra-modal and small inter-class discrepancy in text-based person search, we argue that the prototype can play an important role in this task. 

\begin{figure*}[!t]
    \centering
    \includegraphics[width=7in]{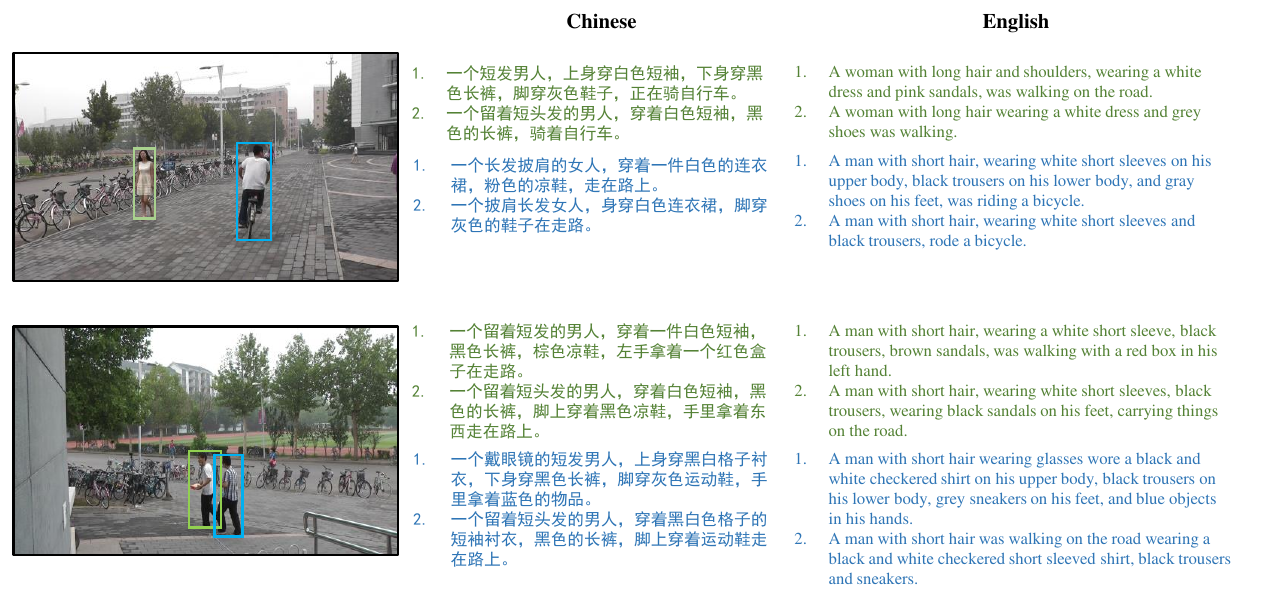}
    \caption{Examples of PRW-TPS-CN dataset. Each bounding box corresponds to an identity and labeled with two sentences. The shown English descriptions are translated from Chinese.}
    \label{Examples of datset}
\end{figure*}

\section{PRW-TPS-CN Dataset}
PRW-TPS-CN is the first Chinese language dataset designed for text-based person search task. It contains rich and structural annotations with open Chinese descriptions. It can robustly evaluate text-based person search methods and help explore language's impact on the task.
\subsection{Dataset Description}
To better evaluate the text-based person search methods, we build a large-scale benchmark dataset based on the PRW\cite{zheng2017person}, termed as PRW-TPS-CN dataset. 
PRW dataset contains the pedestrians' locations in the scene image, thus we can achieve the person detection task. These scene images are the frames captured from the scene videos shot by six cameras, and the main object is the pedestrian.

To build PRW-TPS-CN dataset, the pedestrian bounding boxes are cropped from the scene images. Then, a large number of workers are employed to describe all the prominent person characteristics in the bounding box as long as possible in natural Chinese. Each image is required to be described by two different workers to keep the descriptions less repetitive. Each sentence is demanded to describe a person from top to bottom, which is in line with the natural description order. These ordered descriptions mainly focus on the person appearance (e.g., hair, clothes, pants, accessories), actions, and interactions with objects around. These structured descriptions are more suitable to extract person features and context features. Examples of our proposed dataset are shown in Figure. \ref{Examples of datset}.

\begin{table}[!t]
\begin{center}
\caption{Sentence Length Distribution in PRW-TPS-CN dataset.}
\label{Sentence Length Distribution in PRW-TPS-CN dataset.}
\renewcommand\arraystretch{1.2}
\begin{tabular}{|c|c|c|c|c|c|c|}
\hline
\bf{Length} & \bf{(20,30]} & \bf{(30,40]} & \bf{(40,50]} & \bf{(50,60]} & \bf{(60,70]} & \bf{(70,102]}\\
\hline 
Num & 5,201 & 16,880 & 15,178 & 7,462 & 1,872 & 474 \\
\hline
\end{tabular}
\end{center}
\end{table}

\subsection{Dataset Statistics}
For the PRW-TPS-CN dataset, there are 11,776 scene images including 8,656 described images, 32,206 bounding boxes with 933 valid IDs. There are a total of 47,102 sentences with 2,002,415 Chinese characters, 3,432 Chinese phrases, and 1,300 unique Chinese characters. The longest sentence in our dataset has 107 characters, and the average sentence length is 42.5. The sentence length distribution is shown in Table. \ref{Sentence Length Distribution in PRW-TPS-CN dataset.}. With long sentences, the dataset becomes more informative and can train a more robust network. We also translate the sentences into English, and the average length is 28.79. Both the Chinese and English average sentence length of our dataset is much more than 23.5 words of CUHK-PEDES\cite{li2017person}, 5.18 words of MS-COCO Caption\cite{lin2014microsoft} and 10.45 words of Visual Genome\cite{krishna2017visual}. With these bilingual descriptions, the PRW-TPS-CN dataset can evaluate the methods more comprehensively. 

\begin{figure*}[!htbp]
    \centering
    \includegraphics[width=0.8\linewidth]{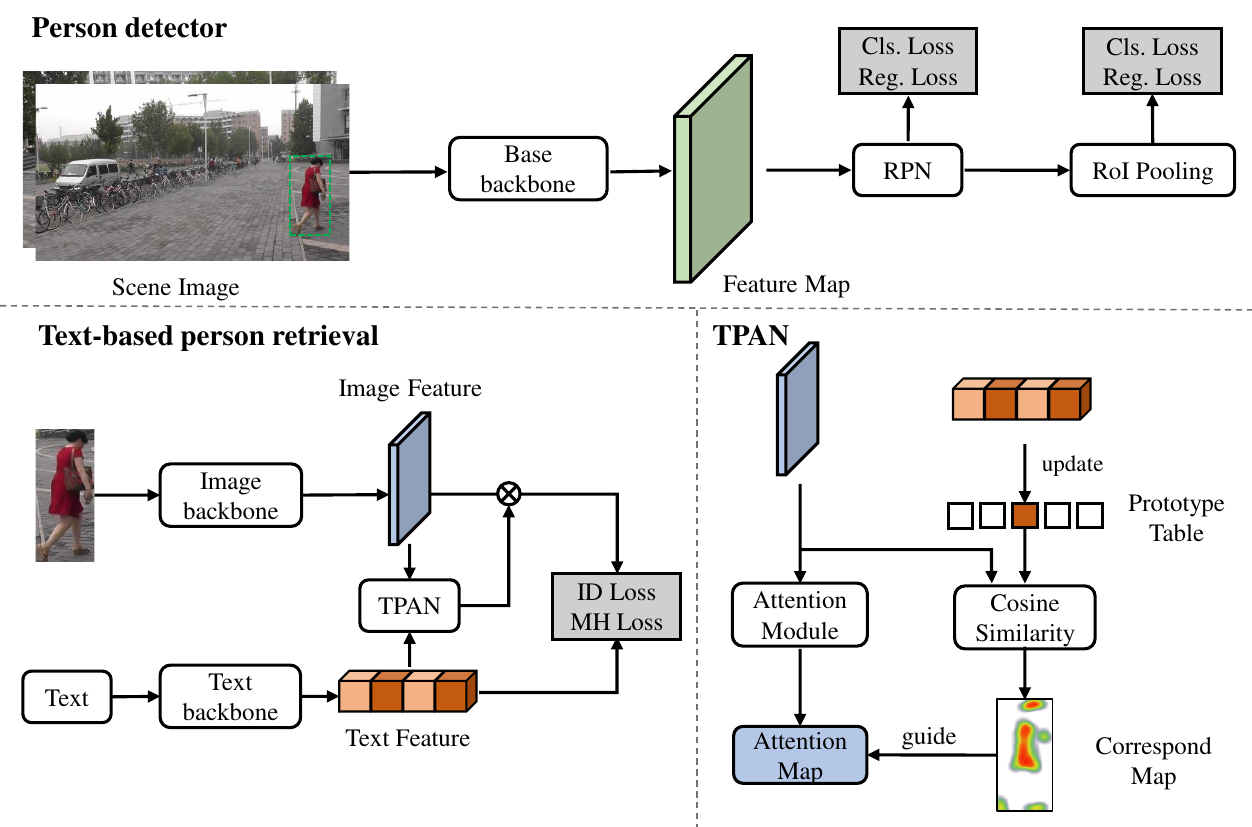}
    \caption{The overall architecture of our text-based person search model. It consists of person detector and text-based person retrieval module and TPAN.}
    \label{framework}
\end{figure*}

\subsection{Evaluation Protocol} \label{Evaluation}
We split the train/test set in a 2:1 ratio of descriptions. We choose 5,664 scene images for training, including 482 person IDs ranging from 1 to 482, 14,896 bounding boxes, and 29,792 descriptions. We randomly select 1/10 scene images and their descriptions to build the validation set. 

For the test set, there are 6,112 scene images and 451 person IDs ranging from 483 to 933. To get the queries, we select 2,992 scene images and describe all the bounding boxes. There are 16,404 ground truth bounding boxes and 17,310 descriptions for query. For the gallery, we use all the 6,112 test scene images.

For testing, we crop bounding boxes from scene images. Given a query text, all detected bounding boxes are calculated with similarity scores. We rank these bounding boxes according to their similarity to the query text.    
Based on the ranking, we employ the widely used Cumulative Matching Characteristic (CMC)\cite{moon2001computational} curve to reflect the search precision.
Considering that a text query has multiple ground truths and CMC is hard to reflect the total ranking results, we also employ the mean Average Precision (mAP) method. mAP considers both search precision and recall and measures the ranking results more comprehensively.

\section{Proposed Network}
In this section, we will describe the proposed network and TPAN in detail. The overall framework is illustrated in Figure. \ref{framework}.

\subsection{Overview}
With the PRW-TPS-CN dataset, there are more and longer text descriptions for each person. We take advantage of the rich texts and propose the TPAN to build the text prototype tables. The prototypes can maintain the salient text features and help guide the generation of image attention.

The network consists of three components: person detector, text-based person retrieval, and TPAN. Given a pair of person text and scene image, the person detector is employed to detect the bounding boxes from images, which is shown at the top of Figure. \ref{framework}. 
For text-based person retrieval, the bounding boxes and texts are fed into the image backbone and text backbone to get features. The image and text features are fed into TPAN to guide the image attention generation. Then, the text features and attended image features are employed to calculate the loss.
For TPAN, the image features are processed with the attention module to get the attention maps, and the text features are used to update the prototype table. The corresponding maps are the cosine similarity of image features and prototype features. The corresponding maps are used to guide the generation of attention maps. The attended image features are the dot product of image features and generated attention maps.

\subsection{Network Components}
Object detectors are usually divided into two categories according to their stages: two-stage and one-stage detectors. Considering the experimental efficiency and final accuracy, we adopt Faster R-CNN as our detector for most experiments. Besides, we also test three representative detectors in this paper: RetinaNet \cite{lin2017focal}, FCOS \cite{tian2019fcos}, YOLO \cite{redmon2018yolov3}.
As for text-based person retrieval, we choose the widely-adopted dual-path structure. We use ResNet\cite{he2016deep} for the image backbone and BiLSTM\cite{hochreiter1997long} as the text backbone. By employing this dual-path structure, TPAN can be well implemented.  

For convenience, in text-based person retrieval, we denote the input image as $i$ and its description as $t$. The extracted image feature map is denoted as $\bf{F_i}\in \mathbb{R}^{C\times H\times W}$, where $C$ is the number of feature channels with height $H$ and width $W$. The extracted text feature is denoted as $\bf{F_t}\in \mathbb{R}^{D\times 1}$, where $D$ is the dimension of the text feature.

\subsection{Text Prototype Attention Network}
Considering alleviating the inconsistency between person detection and text-based person search is crucial for traditional person search task, we propose TPAN to train the text-based person retrieval to redress the detection misalignment. TPAN is illustrated in Figure. \ref{framework}. TPAN combines a person's texts as the prototype to guide the image to generate an attention map that highlights the text-related regions.

In the following content, we first introduce the extraction of image and text features. Based on these features, we show the implementation of TPAN. The overall function is also shown.

\noindent \textbf{Image Path} We employ Squeeze-and-Excitation(SE) \cite{hu2018squeeze} block as the channel and spatial attention module for its lightweight mechanism and well-performance. Given the image feature map $\bf{F_i}$, we reduce its dimension for efficient embedding $\bf{F_c} \in \mathbb{R}^{D\times H\times W}$ and generate 
the channel descriptors $\bf{f_c}\in \mathbb{R}^{D}$ by averaging spatial features. Then, the descriptors are processed by two FC layers and activation functions to produce the attention $\bf{A_c}$. The process is formulated as follows:

\begin{equation}
\label{eq:4.1}
\bf{A_c} = \sigma(\bf{W^{c}_{2}} \delta(\bf{W^{c}_{1}} f_c))
\end{equation}

\begin{equation}
\label{eq:4.2}
\bf{\hat F_c} = \bf{A_c} \odot \bf{F_c}
\end{equation}

\noindent where $\bf{W^{c}_{1}} \in \mathbb{R}^{{\frac{D}{r}}\times{D}}$ and $\bf{W^{c}_{2}} \in \mathbb{R}^{{D}\times \frac{D}{r}}$ are the parameters of the two FC layers. $r$ is the reduction ratio of channels in the SE block. $\sigma$ and $\delta$ are sigmoid and ReLU activation functions. The channel attention feature $\bf{\hat F_c} \in \mathbb{R}^{D\times H \times W}$ is element-wise multiplication of $\bf{A_c}$ and $\bf{F_c}$.

For the spatial attention, $\bf{\hat F_c}$ is channel-wise average pooled and flattened as $\bf{f_s} \in \mathbb{R}^{HW}$. The spatial descriptors are also fed into two FC layers and activation functions. The process is formulated as follows:

\begin{equation}
\label{eq:4.3}
\bf{A'} = \sigma(\bf{W^{s}_{2}} \delta(\bf{W^{s}_{1}} f_s))
\end{equation}
\noindent where $\bf{W^{s}_{1}} \in \mathbb{R}^{{\frac{HW}{r'}}\times{(HW)}}$ and $\bf{W^{s}_{2}} \in \mathbb{R}^{{(HW)}\times \frac{HW}{r'}}$ are the parameters of the two FC layers. $r'$ is the reduction ratio in the SE block. $\bf{A'}\in \mathbb{R}^{HW}$ is reshaped to $\bf{A}\in \mathbb{R}^{H\times W}$ as the final image attention map.

\noindent \textbf{Text Path} In previous text-based person retrieval, building interaction between person image and text is the most common pattern for learning the relationship between image and text. The image and text are matched at the instance level for this pattern. 
However, in text-based person search, due to the limited information in text, it is hard for the model to exploit the relationship between image and text fully.

Thus, similar to that in OIM loss\cite{xiao2017joint}, we build a lookup table to store the prototype text feature of a person. For each identity, the prototype text features can be checked in the lookup table. These prototype text features are used to guide the generation of image attention maps. In detail, we denote the prototype table as $P\in \mathbb{R}^{N\times D}$, where $N$ is the number of person identities. The $k$-th identity in $T$-th iteration of $P$ is updated as follow:
\begin{equation}
\label{eq:4.4}
\bf{P^{T}_{(k)}} = \lambda \bf{P^{T-1}_{(k)}} + (1 - \lambda)\bf{f}^{T}_{t(k)}
\end{equation}
where $\bf{f}_{t(k)}$ is the text features of the person $k$.

In TPAN, we employ text prototypes to guide the generation of image attention maps. 
The target correspond map is generated by cosine similarity of $\bf{P_{(k)}}$ and $\bf{F_{i}}$. The process is calculated as follows:
\begin{equation}
\label{eq:4.5}
\bf{A}^{h,w}_{target(k)} = 
\delta(\frac{\bf{P_{(k)}} \cdot \bf{F}^{h,w}_{i}}{||\bf{P_{(k)}}||_2 \cdot ||\bf{F}^{h,w}_{i}||_2})
\end{equation}
where $\bf{A_{target}}$ is the text-related attention map of image.

With this target response map, we employ MSE as the type of loss function. The loss function can be formulated as follow:
\begin{equation}
\label{eq:4.6}
\mathcal{L}_{guide(k)}=\sum^{H}_{h} \sum^{W}_{w} ||\bf{A^{h,w}_{target(k)}} - \bf{A}^{h,w}_{(k)}||_{2}
\end{equation}
\begin{equation}
\mathcal{L}_{guide} = \frac{1}{N}\sum^{N}_{k=1}\mathcal{L}_{guide(k)}
\end{equation}
where $N$ is the total number of identities. For guiding the attention map $\bf{A}$, the gradient flew to $\bf{A_{target}}$ is set to zero in the back-propagation stage.

The image attention map $\bf{A}$ is guided by the text prototype, containing the overall salient features of the person. Thus, applying the image attention maps to images can make the images focus on the text-related regions. The image attention maps can help extract the images features more consistently, which largely alleviates the detection misalignment.

\subsection{Overall Loss Function}
The whole network includes two training stages. For the person detector, the network is supervised with an objectness loss ($\mathcal{L}_{rpn\_cls}$) and a regression loss ($\mathcal{L}_{rpn\_reg}$) for Region Proposal Network(RPN), and a classification loss(($\mathcal{L}_{cls}$)) and a regression loss($\mathcal{L}_{reg}$) for Region of Interest(RoI) head. The loss function of the person detector is formulated as follows:
\begin{equation}
\label{eq:4.7}
\mathcal{L}_{det}=\mathcal{L}_{rpn\_cls}+\mathcal{L}_{rpn\_reg}+\mathcal{L}_{cls}+\mathcal{L}_{reg}
\end{equation}

For the text-based person search module, we adopt identification loss($\mathcal{L}_{ID}$) to learn the identity feature and Max of Hinge($\mathcal{L}_{MH}$) loss\cite{faghri2017vse++} to embed image and text feature into a common space. With these two loss functions, the text-based person search loss is formulated as follows:
\begin{equation}
\label{eq:4.8}
\mathcal{L}_{tps}=\mathcal{L}_{ID}+\mathcal{L}_{MH}+\mathcal{L}_{guide}
\end{equation}

\section{Experiments}
In this section, we conduct experiments on the proposed benchmark dataset PRW-TPS-CN. We first evaluate some attributes of this dataset. Then, we compare TPAN with other person search methods and report the ablation study results of each component in our proposed TPAN. 

\subsection{Implementation Details}
\noindent \textbf{Network details}
Our proposed network is divided into person detection and text-based person retrieval. Our person detection is implemented based on Faster R-CNN\cite{ren2015faster}, the base network has four blocks from conv1 to conv4 in ResNet-50. Feature Pyramid Network (FPN) is employed to extract person features from multiple scales. For the text-based person retrieval, we use ResNet-50 and BiLSTM as the image and text feature extractor, and the feature dimensions are both 1,024. The reduced ratio $r$ and $r'$ in TPAN are empirically set to 16 and 12. The update ratio $\lambda$ of the prototype table in Eq. (\ref{eq:4.4}) is set to 0.5. All the models are trained in NVIDIA GeForce RTX 3090 GPU and PyTorch framework. 

\noindent \textbf{Model training}
For the person detector, it includes RPN and RoI modules. We input the image to the backbone and use FPN to extract multi-scale features from conv2$\sim$conv5. For training the RPN, we adopt scales {8, 16, 32} and aspect ratios {1, 2, 3}. Non Maximum Suppression (NMS) of 0.7 threshold is used to filter out redundant boxes, and the IoU of the proposals is larger than a threshold of 0.7. RoI Align is employed to extract the features of proposals. We train the detector for 24 epochs using the Stochastic Gradient Descent (SGD) algorithm with momentum set to 0.9. The learning rate is 0.02 initially and decreases to 1/10 at the 16 and 22 epoch. The batch size is set to 4.

For text-based person retrieval module, it includes image and text path. The images are resized to $384\times 128$. We use ResNet-50 to extract image features, and the SE block generates the attention maps from the features after Conv5. For texts, we first use Bert to transfer them into 768-dimension embedding and feed them into BiLSTM to extract text features. We adopt the Adam algorithm with default hyper-parameters in PyTorch, and the network is trained for 30 epochs. The learning rate is set to 0.0001 for image and text path and decreases to 1/10 at 20 epoch. The batch size is set to 64, and we use horizontal flip as data augmentation.

\noindent \textbf{Model Testing}
In the test stage, we first use a trained detector to crop person images from scene images. We keep the detected person images with classification scores of person above 0.5. The detected person images and query texts are fed separately into the image feature extractor and text feature extractor. We calculate the cosine similarity between query texts and person image features and sort the image according to the similarity. For each query text, the sorted images are the retrieved results. As mentioned in section \ref{Evaluation}, we use CMC and mAP as the evaluation protocol.

\begin{table}[!t]
\small
\caption{Impact of text number in PRW-TPS-CN dataset}
\label{text number}
\centering
\renewcommand\arraystretch{1.3}
\begin{tabular}{|c|c|c|c|c|c|c|}
\hline
\bf{Methods} & \bf{mAP} & \bf{Rank-1} & \bf{Rank-5} & \bf{Rank-10}\\
\hline
baseline (One Text) & 23.02 & 39.98 & 62.38 & 70.06 \\
\hline
baseline & 27.71 & 44.90 & 67.12 & 76.29\\
\hline
TPAN (One Text) & 22.10 & 39.20 & 61.80 & 69.88 \\
\hline
TPAN & \bf{29.00} & \bf{46.63} & \bf{68.98} & \bf{75.49} \\
\hline
\end{tabular}
\end{table}

\begin{table}[!t]
\small
\caption{Impact of language in PRW-TPS-CN dataset}
\label{language}
\centering
\renewcommand\arraystretch{1.3}
\begin{tabular}{|c|c|c|c|c|c|c|}
\hline
\bf{Language} & \bf{mAP} & \bf{Rank-1} & \bf{Rank-5} & \bf{Rank-10}\\
\hline
English & 26.80 & 42.90 & 66.77 & 74.30 \\
\hline
Chinese & \bf{29.00} & \bf{46.63} & \bf{68.98} & \bf{75.49} \\
\hline
\end{tabular}
\end{table}

\begin{table}[!t]
\small
\caption{Impact of description order in PRW-TPS-CN dataset}
\label{description order}
\centering
\renewcommand\arraystretch{1.3}
\begin{tabular}{|c|c|c|c|c|c|c|}
\hline
\bf{Dataset} & \bf{mAP} & \bf{Rank-1} & \bf{Rank-5} & \bf{Rank-10}\\
\hline
PRW-TPS-CN(shuffle) & 28.11 & 44.90 & 67.47 & 74.70 \\
\hline
PRW-TPS-CN & \bf{29.00} & \bf{46.63} & \bf{68.98} & \bf{75.49} \\
\hline
\end{tabular}
\end{table}

\subsection{PRW-TPS-CN Dataset Analysis}
In this section, we analyze the necessity of some settings in the PRW-TPS-CN dataset. First, we analyze the necessity that the number of description texts required for a pedestrian is 2. We then compared the effectiveness of different languages and top-to-bottom description order in text-based person search.

\subsubsection{\textbf{Impact of Text Number}} \label{textnumber}
We compare the impact of text number in the training stage with the dataset split in Section \ref{Evaluation}. ``One Text'' means randomly selecting one text of a person image, and there are 5,134 images and 14,896 descriptions for training. The results are shown in Table. \ref{text number}.

As can be seen from the table, while using the baseline model, only one text for training leads to 4.69\%/4.92\% degradation on mAP and Rank-1. While using the TPAN, only one text for training leads to 6.90\%/7.43\% degradation on mAP and Rank-1. Based on all the results, the reduction of text per person image leads the model harder to train, thus the performance decreases much. Therefore, it is reasonable for the PRW-TPS-CN dataset to contain two texts per person, which enables the model to fully learn the relationship between text and image. 
In addition, compared to baseline, TPAN has a higher improvement in two texts than in one text. This result shows that the maintained text feature prototypes keep the salient features of each person and help generate better image attentions, which fully utilizes the text information.   


\begin{table*}[!t]
\small
\caption{Comparison with other methods. The red color means the best result of the text query. The blue color means the best results of the image query.}
\label{SOTA Comparison}
\renewcommand{\arraystretch}{1.3}
\centering
\renewcommand\arraystretch{1.2}
\begin{tabular}{|c|c|c|c|c|c|c|c|}
\hline
\bf{Methods} & \bf{Query Modality} & \bf{Publication} & \bf{Stage} & \bf{mAP} & \bf{Rank-1} & \bf{Rank-5} & \bf{Rank-10}\\
\hline
QEEPS\cite{munjal2019query} & \multirow{4}{*}{Image} & CVPR'19 & \multirow{4}{*}{1} & 37.10 & 76.70 & - & - \\
BINet\cite{dong2020bi} & & CVPR'20 & & 45.30 & 81.70 & - & - \\
NAE\cite{chen2020norm} & & ICCV'20 & & 43.30 & 80.90 & - & - \\
AlignPS\cite{yan2021alignps} & & CVPR'21 & & 45.90 & 81.90 & - & - \\
\hline
CLSA\cite{lan2018person} & \multirow{4}{*}{Image} & ECCV'18 & \multirow{4}{*}{2} & 38.70 & 65.00 & - & - \\
RDLR\cite{han2019re} & & ICCV'19 & & 42.90 & 70.20 & - & - \\
IGPN\cite{dong2020instance} & & CVPR'20 & & \bf{\textcolor{blue}{47.20}} & 87.00 & - & - \\
TCTS\cite{wang2020tcts} & & CVPR'20 & & 46.80 & \bf{\textcolor{blue}{87.50}} & - & - \\
\hline
SDPG\cite{zhang2021text} & Text & arXiv'21 & 1 & 11.93 & 21.63 & 42.54 & 52.99 \\ 
\hline
TPAN & Text & - & 2 & \bf{\textcolor{red}{29.00}} & \bf{\textcolor{red}{46.63}} & \bf{\textcolor{red}{68.98}} & \bf{\textcolor{red}{75.49}} \\
\hline
\end{tabular}
\end{table*}

\subsubsection{\textbf{Impact of Language}}
There are both Chinese and English in PRW-TPS-CN dataset, we compare the impact of these two languages in text-based person search. For a fair comparison, we employ the same preprocessing and training strategy for English. We use these English descriptions for training and testing. The results are shown in Table. \ref{language}.

As can be seen from the table, training in English is slightly inferior to that of Chinese on mAP and Rank-1 by 2.2\%/3.73\%, which can be regarded as a similar performance. The performance degradation may be due to the errors introduced by the machine translation process, the extra preprocessing required by English, or a different training strategy. These results show that the PRW-TPS-CN dataset can also be applied to English-based person search.

\subsubsection{\textbf{Impact of Language Structure}}
To evaluate the effectiveness of describing the person image from top to bottom, we randomly shuffle the phrases in each description. The results are shown in \ref{description order}. From this table, it can be seen that the performance of TPAN in the shuffled PRW-TPS-CN dataset is inferior to that in structured PRW-TPS-CN dataset. The results show that the structured descriptions are helpful to building better connections between different modalities and improving the performance.

\begin{table}[!t]
\begin{center}
\caption{Comparison between PRW-TPS-CN and PRW-TBPS dataset}
\label{Comparison between PRW-TPS-CN and PRW-TBPS dataset}
\renewcommand\arraystretch{1.2}
\begin{tabular}{|c|c|c|}
\hline
\bf{Dataset} & \bf{PRW-TPS-CN} & \bf{PRW-TBPS}\\
\hline
\#sentences & 47,102 & 19,009 \\
\hline
\#annotated images & 8,656 & 5,704 \\
\hline
\#characters(words) & 2,002,415 & 1,318,445 \\
\hline
\#sentence/person & 2 & 1 \\
\hline
\#query & 17,310 & 4,112 \\
\hline
\makecell[c]{average sentence \\ length} & 42.50(28.79) & 24.96 \\
\hline
description criteria & Top to bottom & - \\
\hline
language & Chinese/English & English \\
\hline 
\end{tabular}
\end{center}
\end{table}

\subsection{Dataset Peculiarity}
It is worth noting that another text-based person search dataset named PRW-TBPS has been proposed by \cite{zhang2021text}.
We demonstrate the comparison in Table. \ref{Comparison between PRW-TPS-CN and PRW-TBPS dataset} and summarize the advantages of our dataset: 

1) Adequate data. 
PRW-TPS-CN dataset contains 47,102 sentences and 11,776 scene images with 8,656 described, which is much more than 19,009 and 5,704 in PRW-TBPS. 
Each bounding box was annotated with two descriptions in PRW-TPS-CN but only one description in the train set of PRW-TBPS. We provide 17,310 query texts for testing, which is much more than the query number 4,112 of PRW-TBPS\cite{zhang2021text}, 2,900 of CUHK-SYSU\cite{xiao2017joint}, and 6,148 of CUHK-PEDES\cite{li2017person}. More queries make the evaluation more robust and persuasive. The average sentence length of our dataset is also longer than PRW-TBPS, which means our descriptions are more informative and can train a more robust network. 

2) Structured descriptions. Each sentence describes a person in a bounding box from top to bottom in PRW-TPS-CN, but PRW-TBPS has no specific description criteria. These structured descriptions are more suitable to extract person features and context features. 

3) Bilingual descriptions. We translate the Chinese descriptions into English in PRW-TPS-CN as another version dataset, but PRW-TBPS has only English descriptions. The bilingual descriptions can evaluate the methods more comprehensively.

\subsection{Compared Methods}
Due to that there is no public method designed for text-based person search, we select some state-of-the-art person search methods for comparison. These methods includes one-step methods: QEEPS\cite{munjal2019query}, BINet\cite{dong2020bi}, NAE\cite{chen2020norm}, AlignPS\cite{yan2021alignps}, and two-step methods: CLSA\cite{lan2018person}, RDLR\cite{han2019re},
IGPN\cite{dong2020instance}, TCTS\cite{wang2020tcts}. Besides, we also compare our methods with SDPG\cite{zhang2021text}, a one-step text-based person search method.

Table. \ref{SOTA Comparison} reports the performance of our proposed TPAN and gives comparisons with other traditional image-based person search methods. The main difference between these methods is that SDPG and our TPAN use person descriptions to retrieve the target person, but other methods use person images as queries. The gallery of these methods contains all 6,112 testing images under the benchmarking setting in PRW\cite{zheng2017person}. Besides, it is worth noting that the query amount in PRW-TPS-CN dataset is 17,310, which is much more than 4,114 in other mentioned methods.

TPAN exceeds SDPG in the case of more text queries by 17.07\%/25.00\% on mAP and Rank-1. It proves the superiority of text prototypes for text-based person search, which can effectively highlight the text-related image regions. Text information is more fully utilized.

However, it can be seen that there is a performance gap between traditional image-based person search and text-based person search. AlignPS is superior to TPAN by 16.90\%/35.27\% on mAP and Rank-1. TCTS outperforms TPAN by 17.80\%/40.87\% on mAP and Rank-1. On the one hand, such a performance gap is due to that text contains relatively limited information compared to images and text-based person search involves cross-modal issues. On the other hand, there is a large room for text-based person search to improve.

The above comparisons prove the superiority of TPAN in the text-based person search task and also demonstrate the task-specific inferiority compared with the image-based person search. For TPAN, it largely narrates the performance gap between text-based person search and image-based person search and bridges a possible transition.

\begin{table*}[!t]
\small
\caption{Ablation Study on PRW-TPS-CN dataset}
\label{Ablation}
\centering
\renewcommand\arraystretch{1.3}
\begin{tabular}{|c|c|c|c|c|c|c|}
\hline
\bf{Methods} & \bf{Prototype} & \bf{Instance} & \bf{mAP} & \bf{Rank-1} & \bf{Rank-5} & \bf{Rank-10}\\
\hline
baseline & & & 27.71 & 45.40 & 67.12 & 74.29 \\
\hline
TIAN & & \checkmark & 26.12 & 43.48 & 65.78 & 73.22 \\
\hline
TPAN & \checkmark & & \bf{29.00} & \bf{46.63} & \bf{68.98} & \bf{75.49} \\
\hline
\end{tabular}
\end{table*}

\begin{table*}[!t]
\small
\caption{Impact of different backbones}
\label{backbone}
\centering
\renewcommand\arraystretch{1.3}
\begin{tabular}{|c|c|c|c|c|c|c|}
\hline
\bf{Image Model} & \bf{Language Model} & \bf{mAP} & \bf{Rank-1} & \bf{Rank-5} & \bf{Rank-10}\\
\hline
ResNet-50 & BiLSTM & 29.00 & 46.63 & 68.98 & 75.49 \\
\hline
ResNet-101 & BiLSTM & 29.60 & 47.32 & 69.86 & 76.26 \\
\hline
ResNet-50 & BERT & 29.74 & 47.42 & 69.91 & 76.21 \\
\hline
ResNet-101 & BERT & \bf{30.30} & \bf{47.77} & \bf{70.51} & \bf{76.96} \\
\hline
\end{tabular}
\end{table*}

\begin{table*}[!t]
\small
\caption{Impact of different detection models}
\label{detectors}
\centering
\renewcommand\arraystretch{1.3}
\begin{tabular}{|c|c|c|c|c|c|}
\hline
\bf{Detector} & \bf{Cropped Image Number} & \bf{mAP} & \bf{Rank-1} & \bf{Rank-5} & \bf{Rank-10}\\
\hline
FCOS \cite{tian2019fcos} & 11,892 & \bf{30.75} & 44.64 & 66.79 & 73.50 \\
\hline
RetinaNet \cite{lin2017focal} & 26,789 & 26.79 & 45.17 & 67.49 & 74.41 \\
\hline
YOLOv3 \cite{redmon2018yolov3} & 27,685 & 26.66 & 45.36 & 67.00 & 73.80 \\
\hline
Faster R-CNN \cite{ren2015faster} & 33,576 & 27.76 & \bf{45.50} & \bf{67.70} & \bf{75.11} \\
\hline
\end{tabular}
\end{table*}

\subsection{Ablation Study}
In this section, we first validate the effectiveness of each component of TPAN. Second, we show the impact of different detectors. Subsequently, we replace various backbone networks to observe their impacts on search performance.

\subsubsection{\textbf{Impact of TPAN}} \label{Impact TPAN}
In TPAN, we use prototype-guided attention to generate the image attention map. To validate the impact of the text prototype, we also use the text instance to generate the attention map named Text-Instance Attention Network (TIAN). Different from keeping and updating a prototype lookup table in TPAN, TIAN directly uses the corresponding texts of the person images. The results are shown in Table. \ref{Ablation}.  

From the table, it can be seen that TIGA has worse performance than the baseline on all the evaluation metrics. It is due to that different descriptions of the same person often have different emphases. Thus TIGA may lead images to learn less consistent features. However, TPAN is superior to the baseline by 1.29\%/1.23\% on mAP and Rank-1. This improvement is achieved by keeping the text prototypes. It is due to that text prototypes contain more information about person identities, which largely highlight the text-related regions in image attention maps. Besides, TPAN surpasses TIGA by 2.88\%/3.15\% on mAP and Rank-1, demonstrating the effectiveness of maintaining the text prototypes.

\subsubsection{\textbf{Impact of Text-based Person Retrieval Backbone}}
In Table. \ref{backbone}, we evaluate the performance of the proposed TPAN in different backbones. For the image feature extractor, we select ResNet-50 and ResNet-101 as the backbones. And for the text feature extractor, we choose BiLSTM and BERT as the backbones. For the BERT model, we preprocess the texts into tokens and feed them into BERT to obtain the final 768-dim features. An FC layer is added to embed the text features into 1,024-dim. The initial learning rate is set to 1/10 as that in BiLSTM.

By keeping the text model as BiLSTM, replacing ResNet-50 with ResNet-101 leads to 0.6\%/0.69\% on mAP and Rank-1. Replacing the visual model with ResNet-101 also improves results when the text model is BERT. When the image model is ResNet-50 or ResNet-101, replacing the text model from BiLSTM to Bert also enhances the results. It is concluded that larger models can lead to better search performance. But large models also require more training time.

\subsubsection{\textbf{Impact of Different Detector}}
To evaluate the impact of detectors, we select four detectors including FCOS \cite{tian2019fcos}, RetinaNet \cite{lin2017focal}, YOLOv3 \cite{redmon2018yolov3}, Faster R-CNN \cite{ren2015faster}. We use the pre-trained model in \cite{mmdetection} as the weight of these detectors. These models are pre-trained by MS-COCO\cite{lin2014microsoft} dataset, and the text-based person retrieval employs TPAN. The results are shown in Table. \ref{detectors}.

The table shows that FCOS only detects 11,892 persons, which is less than the ground truth of 16,404 and means that there will be a lot of missing detections. It achieves the best mAP result but the worst Rank-1 Result. The high mAP and low Rank-1 are both due to the missing detections. Thus FCOS is currently not suitable for text-based person search task. 

As for RetinaNet and YOLOv3, they achieve similar results to each other and are better than FCOS. However, Faster R-CNN detects more persons and achieves the best Rank-1 results, which are superior to RetinaNet and YOLOv3 by 0.33\%/0.14\%. Based on the results, Faster R-CNN is more suitable for the current text-based person search. However, the detected persons are about twice the ground truth, which is a problem that text-based person search needs to solve. As in section \ref{Impact TPAN}, the proposed TPAN can address the misalignment in detection.

\begin{figure*}[!t]
    \centering
    \includegraphics[width=1\linewidth]{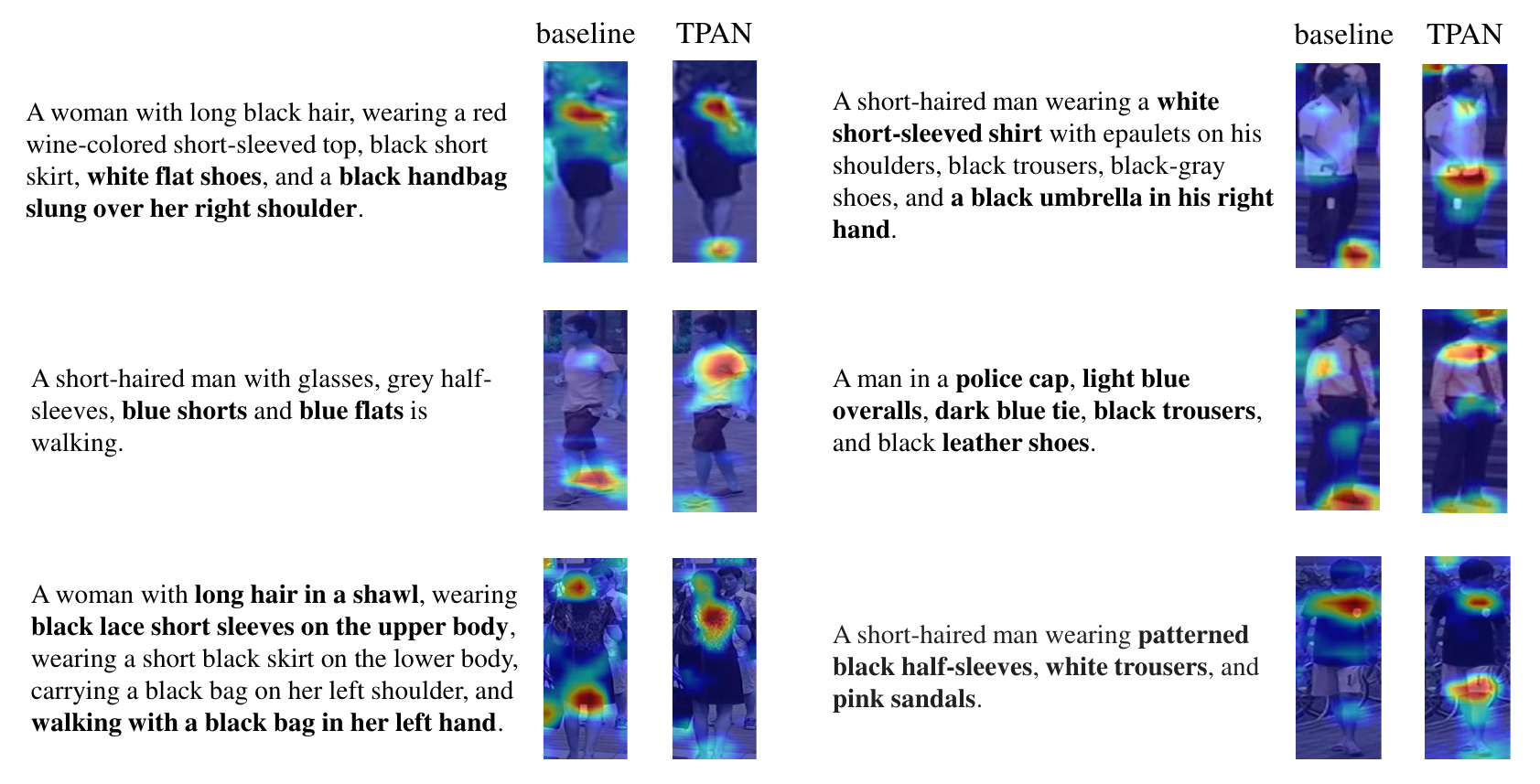}
    \caption{Comparison of baseline and TPAN with respect to Class Activation Maps (CAM) \cite{selvaraju2017grad}. The queries are translated from Chinese into English. Bold words indicate the key information of the text that the image focuses on by using TPAN.}
    \label{CAM}
\end{figure*}

\begin{figure*}[!htbp]
    \centering
    \includegraphics[width=1\linewidth]{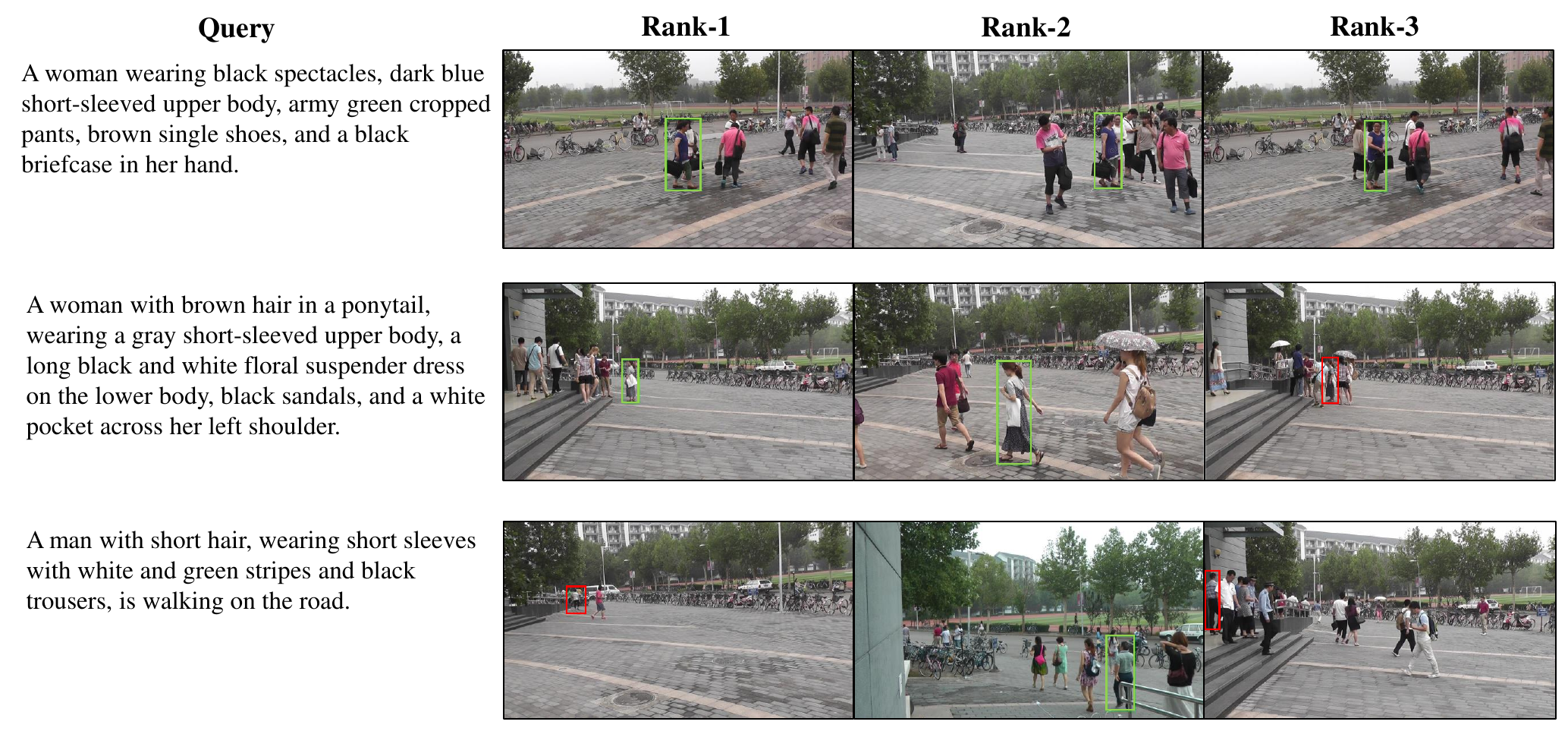}
    \caption{Illustration of the retrieved results. The queries are translated from Chinese into English. The green rectangle represents the correct retrieved person, and the red rectangle represents the incorrect retrieved person.}
    \label{retrieval results}
\end{figure*}

\subsection{Visualization}
\subsubsection{\textbf{Visualization of Retrieval Results}}
To exhibit the effect of TPAN intuitively, we compare the CAM results between baseline and TPAN. The results are demonstrated in Figure. \ref{CAM}. 

Take the first query text from the first row as an example. The baseline model only concentrates on the shoulder of the person, while TPAN pays attention to the shoulder, arm, and shoes of the person simultaneously. For the second text in the first row, the vital feature of the person is the black umbrella held in his right hand, which TPAN can concentrate on, but the baseline model cannot. As can be seen from the figure, the baseline only pays attention to a few regions of person images, which shows that the model pays attention to a few words in the text. It is challenging to achieve text-based person search through these words and regions. However, by employing TPAN, the model can pay attention to more words in the text, resulting in a more comprehensive attention map.

\subsubsection{\textbf{Qualitative Results}}
Figure. \ref{retrieval results} demonstrates some retrieval results of the text-based person search task. The top 3 rows show the cases where the retrieval results are correct: the persons corresponding to the texts appear in the first three ranked scene images. As for the first query, TPAN can use it to accurately search for the target person from scene images containing multiple persons. It shows that our proposed method can effectively extract the text and image features of the person. These features can perform a more accurate search. As for the second query, the size of the target person in the images has significantly changed, but our method is still able to search for her, which shows that our model is robust to person size. The Rank-3 search fails because there are too many occlusions of persons in the image. Besides, for the top 3 queries, the angles of the target persons are different, which shows that our method can adapt to different monitoring angles. 

For the failure case, our approach ranks the target person second. The false retrieval results are due to the person size being too small and the person occlusion area being too large. Nevertheless, the characteristics of persons in false retrieval results are also very similar to the query text, such as the words ``stripes clothes'', ``black trousers'' and ``short hair''.

\section{Conclusion}
In this paper, we focus on the text-based person search task. To solve this problem, we propose a text-based person search dataset named PRW-TPS-CN, in which the persons are described in rich Chinese and translated English descriptions. Considering the existing method for this problem has impractical performance, we propose an intuitive network TPAN to solve this problem. 
In TPAN, the texts of each person are combined as prototypes to guide for generating the image attention maps that highlight the text-related regions, largely eliminating the detection misalignment. To validate the superiority of our method, we conduct extensive experiments on the proposed PRW-TPS-CN dataset. The experimental results demonstrate that our proposed method achieves state-of-the-art performance and vastly outperforms the existing method. Most importantly, we prove the idea of searching for a person in a scene image by text is feasible.

\section{ACKNOWLEDGEMENT}

\newpage
{

}
\end{document}